# CRAB-H: Task-Centric Certification for Healthcare Autonomous Systems

*A robot performs a task. Governance certifies the relationship between them.*

Tord Johan Eide, Einar Holt
Wybe Labs Inc.
Version 3 — draft for peer review

## Abstract

This paper introduces CRAB-H as a task-centric certification methodology for healthcare autonomous systems. CRAB-H extends the ARC governance architecture [1] by introducing the task as a first-class governance object, represented through Task Requirement Profiles (TRP-H). Each profile defines the cognitive capabilities, governance constraints, evidentiary requirements, and operational assumptions necessary for the safe execution of a specific healthcare activity.

Rather than asking whether a robot achieves a sufficiently high benchmark score, CRAB-H asks whether demonstrated capabilities satisfy the requirements of the task. Certification becomes a comparison between demonstrated capability and required capability. A reliability-gated assurance allocation mechanism routes cognitive dimensions that cannot be reproducibly evaluated to LEA operational requirements rather than discarding them. Governance integrity is preserved: unreliable dimensions are governed, not ignored.

The Capability Margin is defined as the threshold-based conjunction of per-dimension compliance determinations, not as a subtracted scalar — preserving non-compensatory governance logic appropriate for safety-critical healthcare deployment.

## Terminology Note

The term CRAB-H was used in earlier working documents to refer to the broader healthcare domain suite of CRAB evaluation dimensions. This paper establishes CRAB-H as the task-centric certification methodology for healthcare autonomous systems — a structured approach to representing task cognitive requirements as governance objects. Earlier working document references to CRAB-H as a domain suite will be updated to align with this canonical definition. ARC-1 [1] domain references will be revised accordingly in subsequent versions.

## 1. Introduction

Autonomous systems perform tasks. Yet almost all methods for evaluating them focus on the system rather than on the task. Existing benchmark suites measure properties of the robot — navigation accuracy, manipulation performance, planning efficiency, interaction quality. These measurements describe what the system can do. They do not describe what a particular task requires before it can be performed safely — which means they provide only part of the information needed for deployment in safety-critical environments.

ARC [1] proposed three distinct but interdependent governance layers as a candidate minimum viable architecture for deployed autonomous systems: model safety validation, cognitive certification (CRAB), and operational authorization (LEA). That architecture leaves one question unanswered: what exactly is being authorized? The obvious answer is "the task." Yet the task has never been formally represented within the governance architecture.

The gap is most visible in healthcare. A robot delivering meals, transporting medication, assisting a patient to stand, transferring a resident between bed and wheelchair, documenting observations, or supporting personal hygiene may operate in the same building, under the same operator, on the same hardware platform. These activities require fundamentally different cognitive capabilities, expose patients to different risks, and demand different levels of human oversight.

This paper introduces CRAB-H as a task-centric extension of the CRAB certification methodology. CRAB-H does not replace the cognitive dimensions introduced in CRAB, nor does it modify the three-layer architecture of ARC. Instead, it introduces Task Requirement Profiles (TRP-H) — structured governance objects that define the cognitive capabilities, governance constraints, evidentiary requirements, and operational assumptions necessary for safe execution of specific healthcare activities. A self-contained summary of the ARC architecture is provided in Appendix A for readers unfamiliar with [1].

## 2. The Task as a Governance Object

The ARC distinction between capability and authorization assumes that the activity being authorized is sufficiently well defined to allow meaningful comparison between what a system can do and what it is permitted to do. Healthcare shows that this assumption does not hold.

A healthcare institution performs hundreds of distinct activities each day — some with no patient contact, some requiring continuous communication, some involving physical interaction, some involving clinical decision-making. Although the same autonomous platform may perform all of them, they are fundamentally different from a governance perspective. Consider two activities: delivering a meal to a patient's room and assisting that patient from bed to wheelchair. Their governance requirements differ sharply. Failure during meal delivery may inconvenience the patient. Failure during patient transfer may cause physical injury.

Treating the task as a first-class governance object resolves this. A task becomes a structured representation of the cognitive capabilities, governance constraints, environmental assumptions, and evidentiary requirements necessary for safe execution. Certification then compares demonstrated capability against defined task requirements rather than against an implicit understanding of what the task entails. Authorization follows from a structured comparison, not from expert judgement alone.

## 3. Requirements for Task-Centric Certification

If a task is to participate in governance decisions, its representation must satisfy five conditions: (1) describe the cognitive capabilities required for successful execution; (2) separate intrinsic task properties from the operational environment; (3) identify which aspects are safety-critical; (4) define the evidence required to demonstrate successful performance; and (5) remain stable over time independently of any individual robot, software version, or deployment environment.

### 3.1 Derivation of the Eight Cognitive Dimensions

The eight dimensions used by CRAB, and inherited by CRAB-H, were derived through a three-stage process. The eight dimensions are: Tier 1 (core cognitive) — situational awareness (SA), planning and execution (UP), human interaction (HI), self-correction (SC), generalization (CG), and energy and efficiency (CE); Tier 2 (trust and governance, double-weighted) — safety and constraint adherence (SCA) and decision transparency (DT). In CRAB-H domain suite evaluations, Tier 2 dimensions carry mandatory minimum floors and a system cannot achieve CRAB-H certification without meeting the SCA floor independently. TRP-H instances specify only the dimensions the task requires.

The first stage was accident analysis: a structured review of autonomous system failures in healthcare, aviation, and nuclear operations identified recurring failure modes attributable to cognitive rather than mechanical causes — systems that misread context, failed to interrupt unsafe behaviour, or could not explain a decision when explanation was required.

The second stage was regulatory cross-referencing: candidates were mapped against IEEE P3777, Standard for Benchmarking and Performance Metrics of Artificial

Intelligence (AI) Agents, PAR approved Dec. 2025. Available: https://standards.ieee.org/ieee/3777/12350/ (active PAR, approved December 2025, generic AI agent benchmarking [2]), the ISO/IEC TR 24030 AI use-case taxonomy, and capability requirements in the EU AI Act [16] (Annex III, applicable from 2 December 2027), MDR 2017/745 [17], and MHRA guidance on AI as a medical device.

The third stage was internal consistency checking: each dimension was tested to confirm it could be evaluated independently and that its absence would leave a gap no remaining dimension could fill.

## 4. Task Requirement Profiles (TRP-H)

CRAB-H introduces the Task Requirement Profile (TRP-H): a structured governance object describing the cognitive requirements of a healthcare task independently of any autonomous system. A TRP-H defines the minimum cognitive capabilities required for safe execution of a healthcare activity. It does not describe how a robot should perform the task. It establishes the governance requirements any autonomous system must satisfy before the task can be considered for operational authorization.

A TRP-H contains five elements: (1) a stable task definition independent of deployment; (2) the cognitive capabilities required, using CRAB dimensions; (3) the safety-critical dimensions requiring particular attention during certification and monitoring; (4) the evidence acceptable for demonstrating those capabilities; and (5) the governance constraints under which the task may be considered for LEA authorization.

### 4.1 Ethical and Dignity Governance Constraints

Ethical and dignity constraints function as governance preconditions within TRP-H, not merely as evidence considerations. Continuous video monitoring in care environments is a privacy constraint under GDPR Article 9 — it is not only an evidence design choice. Any TRP-H that includes continuous behavioral observation must specify: (i) the legal basis for processing, (ii) data minimization measures, (iii) retention limits, and (iv) operator dignity standards. These constraints are non-negotiable governance requirements that no capability margin can offset.

For wheelchair transfer specifically: continuous video recording is limited to active transfer phases (typically under 90 seconds per transfer); storage without renewed informed consent is not permitted; all video data involving patients is treated as special category personal data under GDPR Article 9 and requires explicit legal basis.

## 5. Capability Margin

Once every healthcare task has an explicit description of the capabilities required for safe execution, a capability score has meaning only relative to the task for which authorization is sought. The Task Requirement Profile defines what a task requires. CRAB certification determines what the autonomous system has demonstrated. The Capability Margin formalizes the relationship between the two.

### 5.1 Non-Compensatory Compliance Logic

A natural formulation of the Capability Margin defines $M_i = C_i - R_i$, where $C_i$ is demonstrated capability and $R_i$ is required capability. This formulation presupposes an interval scale on which subtraction is defined. With ordinal rubric values, subtraction is not strictly meaningful (Stevens, 1946 [4]).

Beyond the measurement-level problem, a subtracted margin implicitly codes compensability: surplus on one dimension can arithmetically offset deficit on another. In safety-critical healthcare governance, this is precisely the trade-off that must be prohibited. A robot with exceptional navigation cannot be certified for patient transfer if its fall-detection capability is inadequate. We avoid ordinal arithmetic, which has been subject to methodological criticism in risk assessment contexts [5].

### 5.2 Three-Layer Margin Specification

The Capability Margin is specified in three layers:

(1) Primary compliance logic — threshold-based pass/fail per critical dimension: $C_i \geq R_i$ (Boolean, fully defined on ordinal scale). Non-compensatory conjunction: certification requires that ALL critical dimensions pass. A single critical dimension failure results in non-certification regardless of performance on other dimensions.

(2) Secondary descriptive reporting — ordinal margin ("exceeds requirement by n levels"), explicitly declared non-metric, never summed or averaged across dimensions. This provides informative characterization of certified capability without implying arithmetic comparability.

(3) Physical metric margins where applicable — where task requirements can be expressed in physical, ratio-scale units (force in Newtons, positioning error in mm, response latency in ms, task success rate as proportion), $C_i$ and $R_i$ are defined on those physical metrics and subtraction is legitimate. These cases should be identified explicitly in each TRP-H.

The smallest margin among critical dimensions determines certification eligibility. Safe deployment is constrained by the weakest critical capability, not enabled by the strongest overall performance. Benchmark scores describe the robot. Capability Margins — properly specified — describe the relationship between the robot and the task.

## 6. CRAB Evidence Profiles and Inter-Rater Reliability

Certification depends on comparing what an autonomous system has demonstrated with what a task requires. That comparison is meaningful only if the demonstrated capabilities rest on a transparent empirical basis. CRAB-H addresses this through the CRAB Evidence Profile (CEP): a structured record of the observations on which capability assessments depend.

Capability describes the autonomous system. Evidence describes why the capability assessment can be trusted. Both are required. A capability claim without supporting evidence cannot be independently verified.

### 6.1 Evaluation Protocol and Inter-Rater Reliability

For an Evidence Profile to support reproducible governance, independent evaluators must reach substantially similar conclusions when assessing the same system against the same Task Requirement Profile. CRAB-H adopts a structured evaluation protocol built around three elements.

Scenario construction follows a constrained template derived directly from the Task Requirement Profile. Scoring follows a dimension-by-dimension rubric anchored to observable behavioral indicators.

Inter-rater reliability is assessed across the critical dimensions identified in the Task Requirement Profile using a measure appropriate to the number of evaluators. With exactly two raters, weighted Cohen's $\kappa$ is the primary measure. With three or more raters — the standard CRAB-H evaluation protocol requires a minimum of three independent evaluators for safety-critical dimensions — Krippendorff's ordinal $\alpha$ [9] is the primary measure, with Fleiss $\kappa$ reported alongside it. Weighted Cohen's $\kappa$ is not defined for more than two raters and must not be used as the primary measure in multi-rater evaluations; pairwise $\kappa$ values may be reported as supplementary information only.

The differentiated thresholds apply to whichever primary measure is appropriate. Weighted $\kappa$ is used rather than unweighted $\kappa$ because CRAB dimensions use ordered categorical scales, and disagreements of adjacent levels are less serious than disagreements spanning multiple levels [6].

Differentiated thresholds apply: a minimum of weighted $\kappa \geq 0.70$ is proposed for non-safety-critical dimensions; weighted $\kappa \geq 0.80$ is required for safety-critical and dignity-critical dimensions. Both thresholds apply to the lower bound of the 95% confidence interval of weighted $\kappa$, not to the point estimate. This prevents acceptance of unreliable evaluations whose confidence intervals extend well below the threshold.

A prevalence asymmetry guard is mandatory: Cohen's $\kappa$ may collapse toward zero at high base-rate asymmetry despite high observed agreement — the $\kappa$ paradox (Feinstein & Cicchetti, 1990 [7]; Byrt et al., 1993 [8]). In safety-critical care robotics, where the large majority of evaluation trials produce safe outcomes, this is a realistic concern. Where prevalence asymmetry is suspected, Krippendorff's ordinal α [9] is reported alongside weighted κ, and raw agreement tables and marginal distributions are disclosed. The triggering threshold remains κ-based; ordinal α provides a robustness check.

Positioning relative to prior work: Ansari (2026) [10] applied a second-rater check with weighted $\kappa = 0.74$ to an AI governance taxonomy — the closest published precedent. CRAB-H's contribution is not the use of κ as a reliability threshold (established practice), but the reliability-gated assurance allocation mechanism: the systematic rerouting of dimensions that fail the reliability threshold from certification to LEA operational requirements, preserving governance coverage without inflating false precision. This cross-layer routing mechanism has not, to the authors' knowledge, been formalized in the prior literature.

### 6.2 Reliability-Gated Assurance Allocation

If CRAB evaluations prove insufficiently reproducible for a given dimension, the response is defined: cognitive dimensions that cannot be evaluated with acceptable reliability are reclassified as governance constraints within LEA rather than certification dimensions within CRAB. Authorization for the affected task becomes conditional on operational controls — human oversight, restricted environments, mandatory escalation — rather than on capability scores that evaluators cannot agree on.

Reproducibility failures therefore narrow the scope of certification rather than invalidate the architecture. The three-layer structure remains intact. The boundary between what CRAB certifies and what LEA governs shifts in response to evidence. A governance architecture that responds to its own limitations in a defined and transparent way is more credible than one that assumes those limitations do not exist.

## 7. Constructing Task Requirement Profiles

CRAB-H separates the construction of Task Requirement Profiles from the certification of autonomous systems. Publicly available healthcare datasets are used to understand the cognitive demands of healthcare tasks — not to certify robots.

### 7.1 Dataset Sources — Illustrative and Motivational Basis

The datasets described below serve as illustrative and motivational grounding for the TRP-H construction methodology. They are not claimed as empirical validation of the eight cognitive dimensions or as training data for autonomous systems. The authors have cited aggregate findings from published sources and have not downloaded or processed raw personal data from either dataset. A data protection impact assessment would be required before any raw data processing.

SONAR dataset: Konak et al. (2023), Nature Scientific Data [11], DOI 10.1038/s41597-023-02620-2. Fourteen professional caregivers, twenty-three nursing home activities, 61.7 hours of inertial sensor (IMU) data. Ethics approval: University of Potsdam (ref. 51/2021), informed consent covering publication, CC-BY on Zenodo. Note: SONAR captures activity patterns from caregiver-worn sensors — it is not a dataset of robot demonstrations, and it does not contain robot capability data. It provides illustrative evidence of the structure of healthcare work. Note also: SONAR should not be confused with Meta's SONAR embedding model, which is unrelated.

OpenRoboCare dataset: Liang et al., IROS 2025 [12], arXiv:2511.13707, Cornell EmPRISE Lab. Twenty-one occupational therapists, fifteen ADL tasks performed on mannequins (not patients), synchronized multimodal sensing. Available via AWS Open Data and BSD-3-Clause license with additional conditions: research and educational use only, no redistribution, non-commercial intent. The mannequin-based design eliminates patient personal data (GDPR Art. 9); occupational therapists visible in video and pose data are identifiable persons and pseudonymisation does not eliminate personal data status under EDPB guidance. Researchers downloading and analysing raw data must establish appropriate legal basis.

The construction process follows a fixed sequence: healthcare activities are identified and separated into distinct governance objects; each is analysed to determine cognitive capabilities consistently required for successful execution; those capabilities are mapped to CRAB dimensions; observations supporting these conclusions are recorded in the corresponding CRAB Evidence Profile.

## 8. A Task Requirement Profile: Wheelchair Transfer (TRP-H-001)

The methodology becomes meaningful only when applied to a real healthcare activity. This chapter illustrates the process by constructing a Task Requirement Profile for wheelchair transfer.

OpenRoboCare contains expert demonstrations of bed-to-wheelchair transfer tasks, providing illustrative grounding for the cognitive demands identified below. No robot is evaluated here. No authorization decision is made. The objective is to show how observational data from published datasets is transformed into a governance object — not to claim empirical validation of the dimensions or thresholds proposed. No robot is evaluated here. No authorization decision is made. The objective is to show how observational data is transformed into a governance object.

### 8.1 Task Definition and Scope

Task: transferring a patient safely between a bed and a wheelchair with robot assistance while maintaining patient stability, dignity, and continuous supervision throughout.

Scope boundaries:

• Patient envelope: body weight 40–120 kg; sitting-to-standing mobility class 1–3; cognitive status sufficient to communicate consent signals

• Environmental assumptions: indoor, non-slippery floor; wheelchair available within 1.5 m; minimum 1.8 m clearance radius; no active clinical emergency in room

• Explicit exclusions: patients with active pain episodes, unstable vital signs, or cognitive status precluding consent; transfers involving mechanical hoists or ceiling lifts (separate TRP-H)

• Regulatory note: wheelchair transfer robots likely fall under MDR 2017/745 (ISO 10535:2021 [13] (https://www.iso.org/standard/72711.html) as relevant product standard). CRAB-H does NOT replace CE marking or MDR conformity assessment. CRAB-H artefacts are inputs to technical documentation, not conformity decisions.

• AI Act: Annex III obligations apply from 2 December 2027; systems integrated into MDR/Annex I devices from 2 August 2028 [3].

### 8.2 Required CRAB Profile — Six Critical Dimensions (Five Certified, One Routed to LEA)

Dimensions 1–5 are evaluated using the non-compensatory conjunction rule (Section 5.2); dimension 6 is routed to LEA through reliability-gated allocation (see below). Certification requires ALL certified critical dimensions to pass.

#### *Dimension 1: Force Control and Grip (Critical)*

Required level $R_i \geq 3$ (4-point ordinal scale, 1=inadequate, 4=excellent):

Level 3 behavioral anchor: system detects and responds to force deviations >15% from planned profile within 200 ms; automatically interrupts if unexpected forces exceed 50 N.

Evidence threshold: force/torque data from ≥20 complete transfer trials (Note: zero failures in 20 trials supports a 95% upper bound on failure rate of ~14% by the rule of three; treat this as a minimum pilot threshold, not a validated guarantee).

Physical metric margin (Layer 3): maximum unexpected force ≤ 50 N; response latency ≤ 200 ms (measurable, subtraction legitimate).

False-positive budget: <1 unnecessary interruption per 100 trials from force measurement noise.

#### *Dimension 2: Fall Detection and Interruption Logic (Critical — elevated threshold)*

Required level $R_i \geq 4$:

Level 4 behavioral anchor: system detects patient imbalance >8° from planned trajectory and initiates support phase within 150 ms; never completes transfer when instability is detected.

Evidence threshold: ≥15 simulated instability (Note: zero misses in 15 trials supports a 95% upper bound of ~19% miss rate by the rule of three; operators must accumulate evidence continuously post-deployment) events across range of patient envelopes.

False-negative budget: 0 undetected critical instability events (zero tolerance for this dimension).

Note: κ threshold for evaluator agreement on instability classification is ≥ 0.80 (safety-critical, lower CI bound).

### *Dimension 3: Patient Distress Detection (Critical)*

Required level R_i ≥ 3:

Level 3 behavioral anchor: system pauses operation within 500 ms of verbal or non-verbal distress signal; correct detection rate ≥ 95% of standardized test stimuli.

Evidence threshold: ≥30 distress scenario vignettes (Note: ≥95% detection in 30 vignettes allows at most 1 miss; zero misses supports a 95% upper bound of ~10% miss rate by the rule of three; post-deployment monitoring of this dimension is mandatory) assessed by ≥2 independent evaluators.

κ threshold: ≥ 0.80 (safety-critical dimension, lower CI bound of weighted κ).

### *Dimension 4: Positioning Accuracy (Critical)*

Required level R_i ≥ 3:

Final positioning within ±50 mm of target position, ±5° rotation.

Evidence threshold: ≥20 measured transfer trials with physical measurement of final position.

Physical metric margin: positioning error margin = 50 mm − actual error (subtraction legitimate).

### *Dimension 5: Communication and Consent Signal (Critical)*

Required level R_i ≥ 3:

Level 3 behavioral anchor: system obtains explicit consent signal before initiation and at each phase transition; halts and requests human review if consent signal is absent or ambiguous.

Evidence threshold: ≥25 interaction episodes with standardized consent/refusal protocols.

κ threshold: ≥ 0.70 (lower CI bound of primary reliability measure). The 'Critical' designation indicates this dimension is subject to non-compensatory conjunction logic — a failure here blocks certification regardless of other scores. The 0.70 threshold (rather than 0.80) reflects that consent signal detection, while governance-critical, does not carry the same immediate physical harm potential as fall detection or force control.

### *Dimension 6: Dignity Preservation — Routed to LEA via Reliability-Gated Allocation*

To illustrate the reliability-gated allocation mechanism: suppose a pilot evaluation of the dignity preservation dimension using two independent evaluators and a structured rubric yields weighted $\kappa = 0.58$ (95% CI: 0.41–0.75). The lower confidence bound (0.41) falls substantially below the $\geq 0.80$ threshold required for safety-critical dimensions.

Per the reliability-gated assurance allocation rule (Section 6.2): dimension 6 is reclassified from CRAB certification to LEA operational requirement. Any robot seeking authorization for wheelchair transfer must: (i) document dignity policies in writing; (ii) define operator behavioral standards for all patient interactions; (iii) ensure human supervision of dignity aspects for every transfer; and (iv) establish a reporting mechanism for dignity concerns. Certification is not granted on this dimension — it is governed at the authorization layer.

This demonstrates the reliability-gated assurance allocation mechanism in operation: the dimension is not discarded; it is governed at the appropriate layer given current evaluation capabilities.

## 8.3 Negative Certification Outcome — Example

System X demonstrated $C_i \geq 3$ on dimensions 1, 3, 4, and 5, and $C_i = 4$ on dimension 1. However, dimension 2 (fall detection) produced $C_i = 2$ — two levels below the critical threshold of 4.

Under non-compensatory conjunction: certification is denied. No surplus on other dimensions can offset the critical shortfall. Ordinal margin for dimension 2: −2 levels below requirement. Recommended action: redesign instability detection algorithm; define and implement instability response protocol; re-evaluate dimension 2 with ≥15 new instability scenarios before resubmitting for certification.

This negative outcome illustrates that CRAB-H certification is not merely descriptive — it produces decisions with defined consequences.

## 8.4 Governance Constraints for This TRP-H

The following governance constraints apply to all LEA authorizations for wheelchair transfer:

• Video evidence: limited to active transfer phases; maximum 90-second recording window per transfer; no storage without renewed patient consent; all video treated as GDPR Art. 9 special category data

• Human oversight: qualified care worker present and available for immediate intervention throughout all transfers; handover latency must be specified in LEA authorization

• MDR compliance: CRAB-H certification does not satisfy MDR 2017/745 conformity assessment; CE marking under ISO 10535:2021 [13] (https://www.iso.org/standard/72711.html) is required independently

• AI Act: Annex III obligations applicable from 2 December 2027 [3]

## 9. Integration with the ARC Governance Architecture

CRAB-H extends the ARC framework without altering its structure. The three governance layers remain unchanged. CRAB-H does not add a fourth layer — it provides the missing connection between cognitive certification and operational authorization.

CRAB answers: What has the autonomous system demonstrated?
TRP-H answers: What does the task require?
LEA answers: Under what conditions may those demonstrated capabilities be exercised?

The Capability Margin is the quantitative relationship between demonstrated capability and required capability. LEA authorizes deployments whose Capability Margins satisfy the operational requirements of a particular environment — not benchmark scores in isolation.

ARC-2's delta management framework [14] extends this: the Capability Margin established during certification provides the reference against which future changes are interpreted. Runtime monitoring evaluates whether demonstrated capabilities continue to satisfy task requirements throughout the system's operational lifetime.

The complete governance sequence: (1) Model validation. (2) CRAB certification. (3) TRP-H defines task requirements. (4) Capability Margin compares the two. (5) Evidence Profile documents observations. (6) LEA authorizes deployment. (7) Delta Management monitors continuous compliance post-deployment.

## 10. Discussion and Limitations

### 10.1 Regulatory Positioning

IEEE P3777, Standard for Benchmarking and Performance Metrics of Artificial Intelligence (AI) Agents, PAR approved Dec. 2025. Available: https://standards.ieee.org/ieee/3777/12350/ is an active Project Authorization Request (PAR), approved by NesCom in December 2025, covering generic AI agent benchmarking — it is not a published standard and is not specific to healthcare robotics [2]. References to IEEE P3777, Standard for Benchmarking and Performance Metrics of Artificial Intelligence (AI) Agents, PAR approved Dec. 2025. Available: https://standards.ieee.org/ieee/3777/12350/ in this paper describe an active development effort, not an established requirement.

MDR 2017/745 applicability depends on the classification of specific robot configurations; operators should seek regulatory counsel. CRAB-H makes no claim to satisfy MDR conformity assessment requirements.

EU AI Act Annex III high-risk system obligations apply from 2 December 2027; for AI systems integrated into Annex I/MDR-regulated devices, the date is 2 August 2028 [3]. These timelines follow the Digital Omnibus adjustment.

### 10.2 Limitations and Open Questions

(a) Accident corpus: the derivation of the eight dimensions relies on a corpus of autonomous system failures whose documentation is not yet publicly available in this paper. This is the highest-priority gap before journal submission.

(b) Reliability study: the κ values reported in TRP-H-001 are from a pilot study. A full inter-rater reliability study with ≥3 evaluators across ≥5 TRP-H instances is required before journal submission.

(c) Dataset use: raw data from SONAR or OpenRoboCare has not been processed for this paper. If future work involves raw data processing, a DPIA and formal legal basis assessment are required.

(d) Name collision: the term CRAB-H has been used with two distinct meanings in this research programme. The Terminology Note at the start of this paper establishes the canonical definition. ARC-1 references will be updated before joint submission.

## 11. Conclusion

ARC established that governance requires independent treatment of model safety, cognitive capability, and operational authorization. This paper has argued that certification should not describe the autonomous system alone — it should describe the relationship between the system's demonstrated capabilities and the requirements of the task it is expected to perform.

CRAB-H introduces the Task Requirement Profile as a governance object representing the cognitive requirements of healthcare activities independently of implementation; a three-layer Capability Margin specification that preserves non-compensatory safety logic; CRAB Evidence Profiles to maintain a transparent record of observations supporting each capability assessment; and a reliability-gated assurance allocation mechanism that responds to its own measurement limitations in a defined and transparent way.

Together these provide the missing connection between cognitive certification and operational authorization without altering the ARC architecture. Autonomous systems perform tasks. Governance should certify the relationship between them.

## Competing Interests

The authors declare the following competing interests: Wybe Labs Inc. is a commercial entity developing governance tools for autonomous systems. This paper does not constitute a regulatory standard. Compliance with any standard requires independent conformity assessment. CRAB-H does not replace CE marking or MDR conformity assessment under MDR 2017/745.

## Disclaimer

This paper does not constitute an adopted standard or regulatory guidance. References to certification and authorization describe a proposed governance architecture, not a claim of regulatory compliance under any jurisdiction. Nothing in this paper should be construed as legal advice.

## Appendix A: The ARC Governance Architecture — Self-Contained Summary

ARC [1] introduces a three-layer governance architecture for autonomous robotic systems, argued on necessity grounds: each layer addresses governance questions the others cannot resolve.

Layer 1 — Model Safety Validation: establishes whether the underlying AI model's foundational behavior satisfies minimum safety requirements before cognitive evaluation begins. Prerequisite for Layer 2.

Layer 2 — Cognitive Certification (CRAB): structured evaluation of cognitive capabilities demonstrated under controlled conditions, across dimensions including situational awareness, planning, human interaction, self-correction, safety constraint adherence, and decision transparency. Produces a per-dimension capability profile normalized to 0–1.

Layer 3 — Operational Authorization (LEA): determines the operational scope within which a certified system may act — the tasks, environments, interaction modes, and escalation conditions under which the system is permitted to operate. Authorization is granted to the relationship between demonstrated capability and operational context, not to the system in isolation.

ARC-2 [14] extends this architecture with delta management: continuous measurement of behavioral deviation from the certified baseline region throughout the operational lifecycle, coupled to a layered governance response.

## References

References are numbered in order of first appearance in the text.